\pdfoutput=1
\documentclass{acmart}

\newif\iffinal
\finaltrue 

\usepackage[utf8]{inputenc} 
\usepackage[T1]{fontenc}    
\usepackage{hyperref}
\usepackage{url}            
\usepackage{booktabs}       
\usepackage{amsfonts}       
\usepackage{nicefrac}       
\usepackage{microtype}      
\usepackage{amsmath}
\usepackage{amsthm}
\usepackage{graphicx}
\usepackage{color}
\usepackage{xcolor}                   
\usepackage{framed}
\usepackage{enumitem}
\usepackage{xspace}
\usepackage[ruled, noend, linesnumbered]{algorithm2e}
\usepackage{multirow}

\newcommand{\mynote}[2]
    {{\color{red} \fbox{\bfseries\sffamily\scriptsize#1}
    {\small$\blacktriangleright$\textsf{\emph{#2}}$\blacktriangleleft$}}~}

\newcommand{\todo}[1]{\mynote{TODO}{#1}}

\newcommand{\x}{\textit{\textbf{x\xspace}}}

\renewcommand{\paragraph}[1]{\vspace*{0.2mm}\noindent\textbf{#1}\hspace{6pt}}

\newcommand{\figspc}{}

\title{Prediction Sensitivity: Continual Audit of Counterfactual Fairness in Deployed Classifiers}

\iffinal
\author{Krystal Maughan$^*$}
\email{Krystal.Maughan@uvm.edu}
\affiliation{%
  \institution{University of Vermont}
  \country{USA}
}

\author{Ivoline C. Ngong$^*$}
\email{kngongiv@uvm.edu}
\affiliation{%
  \institution{University of Vermont}
  \country{USA}
}

\author{Joseph P. Near}
\email{jnear@uvm.edu}
\affiliation{%
  \institution{University of Vermont}
  \country{USA}
}
\fi

\begin{abstract}
  As AI-based systems increasingly impact many areas of our lives,
  auditing these systems for fairness is an increasingly high-stakes
  problem. Traditional group fairness metrics can miss discrimination
  against individuals and are difficult to apply after deployment.
  Counterfactual fairness describes an individualized notion of
  fairness, but is even more challenging to evaluate after deployment.
  We present \emph{prediction sensitivity}, an approach for
  continual audit of counterfactual fairness in deployed classifiers.
  Prediction sensitivity helps answer the question: \emph{would
    \textbf{this} prediction have been different, if \textbf{this}
    individual had belonged to a different demographic group}---for
  \emph{every} prediction made by the deployed model. Prediction
  sensitivity can leverage correlations between protected status and
  other features, and does not require protected status information at
  prediction time. Our empirical results demonstrate that prediction
  sensitivity is effective for detecting violations of counterfactual
  fairness.
\end{abstract}

\begin{document}

\maketitle

\section{Introduction}
\renewcommand*{\thefootnote}{\fnsymbol{footnote}}
\footnote{The first two authors contributed equally.}
As AI-based systems take increasingly larger roles in high-stakes
decisions, ensuring fairness in these systems represents a growing
concern.
Previous work has examined the questions of how to audit data and
classifiers for bias before they are
deployed~\cite{buolamwini2018gender, DBLP:conf/fat/MitchellWZBVHSR19},
and how to mitigate bias either during training or as a
post-processing step (e.g.~\cite{calmon2017optimized,
  celis2019classification, zemel2013learning, hardt2016equality}).

Many existing approaches use \emph{metrics} that offer formal
processes for ``measuring'' fairness. \emph{Group fairness}
metrics~\cite{feldman2015certifying, hardt2016equality} measure
disparate treatment of groups in aggregate. These metrics are useful
to demonstrate unfairness, but previous work has shown that group-fair
classifiers can still make clearly unfair predictions for
individuals~\cite{LiptonMC18}. \emph{Individual fairness}
metrics~\cite{dwork2012fairness} require that similar individuals be
treated the same, but the difficulty of formalizing similarity metrics
makes the framework challenging to apply.

\emph{Counterfactual fairness}~\cite{kusner2017counterfactual}
requires that a fair classifier \emph{would have made the same
  prediction} if the individual had belonged to a different
demographic group. Demonstrating that a classifier satisfies
counterfactual fairness typically requires causal information about
the underlying data distribution~\cite{kusner2017counterfactual,
  kaushik2020explaining, kaushik2019learning}, which presents a major
challenge in deploying practical audit systems based on counterfactual
fairness. An important exception is the recent
FlipTest~\cite{black2020fliptest} system, which uses an optimal
transport mapping (rather than causal information) to demonstrate
counterfactual fairness.
Like other existing techniques, FlipTest is designed for auditing
models \emph{before} they are deployed. Far less work has examined
methods for ensuring that classifiers continue to perform well after
deployment---even though novel examples encountered after deployment
are likely to reveal bias in trained
classifiers~\cite{buolamwini2018gender, googlegorillas}.

This paper proposes a new approach called \textbf{prediction
  sensitivity} for \emph{continual audit} of counterfactual fairness
in \emph{deployed classifiers}. In contrast to FlipTest, which
requires the generation and evaluation of explicit counterfactual
testing data during a distinct audit phase of model development (i.e.
before the model is deployed), prediction sensitivity helps answer the
question: \emph{would \textbf{this} prediction have been different, if
  \textbf{this} individual had belonged to a different demographic
  group}---for \emph{every} prediction made by the deployed model.

This kind of continual audit allows deployed systems to raise an alarm
before making a potentially unfair prediction, \emph{even for novel
  examples}. This capability represents an important difference from
previous work, which may miss sources of unfairness due to missing
training data.
Prediction sensitivity, by contrast, will be able to catch the resulting
discrimination when it occurs and is visible---\emph{after} the model
is deployed.

Prediction sensitivity uses the \emph{gradient of the model's
  prediction} with respect to the input to approximate how the
prediction would have changed if specific parts of the input had been
different. Specifically, our approach involves training a
\emph{protected status model} capable of weighting input features
according to their importance for determining the protected status, and
then weighting elements of the gradient using this information. Moreover, the protected status model captures correlations between the protected attribute and other input features, which is significant, since previous work has shown removing the protected attribute~\cite{dressel2018accuracy} or just flipping~\cite{black2020fliptest} its value can still cause discrimination in the model's predictions. We require knowledge of the protected status of individuals at training time (to train the protected status model), but not at prediction time.

Prediction sensitivity can help avoid making biased predictions and
can notify system maintainers of problems with their models. As a
metric for counterfactual fairness, it can be automatically and
efficiently calculated for each prediction and compared against a
threshold set by the system maintainer. The protected status of
individuals (i.e. the protected attribute) does not need to be present
at prediction time in order to calculate prediction sensitivity.

We evaluate prediction sensitivity on both synthetic and real data,
using experiments inspired by FlipTest~\cite{black2020fliptest}. We
construct \emph{match sets} (similar to FlipTest's
\emph{flipsets})---examples for which a trained classifier would have
made the same prediction if the protected status had been different.
Non-members of the match set represent failures of counterfactual
fairness. Our results suggest that prediction sensitivity is
effective at detecting these failures.

\paragraph{Contributions.}
In summary, we make the following contributions:
\begin{itemize}[topsep=0pt]
\item We propose \emph{prediction sensitivity}, a gradient-based
  method for measuring counterfactual fairness
\item We show how to use prediction sensitivity to detect biased
  predictions at the individual level in deployed models
\item We present experimental results suggesting that prediction
  sensitivity is effective for detecting biased predictions
\end{itemize}

\section{Background \& Related Work}

\paragraph{Deep learning.}
In this paper, we focus on machine learning models represented by \emph{artificial neural networks}~\cite{goodfellow2016deep}. A model $\mathcal{F}$ is parameterized by a set of \emph{weights} which are optimized during training; we write $\mathcal{F}(\x)$ to represent a \emph{prediction} made by the trained model on an example $\x$. Deep learning models are typically trained by optimizing a \emph{loss function} $\mathcal{L}$ by calculating the \emph{gradient} of the loss with respect to the weights and updating the model accordingly.


\paragraph{Fairness in machine learning.}
The bulk of previous work on fairness in machine learning attempts to improve \emph{group fairness metrics} at training time, often by the introduction of new kinds of regularization~\cite{calders2009building, woodworth2017learning, zafar2015fairness, zafar2017fairness, agarwal2018reductions, russell2017worlds, celis2019classification, beutel2017data, shankar2017no, zhang2018mitigating, wadsworth2018achieving, celis2019improved, zemel2013learning, louizos2015variational, lum2016statistical, adler2018auditing, calmon2017optimized, feldman2015computational, hardt2016equality}. Many of these approaches are suitable for deep learning, and have been empirically validated using the metrics described above.
These approaches typically apply when each example's features include a \emph{protected attribute} $z \in x$ to indicate the example's \emph{protected status} (as a member or non-member of a protected class).
%
Existing approaches focus on notions of group fairness, and are validated using metrics for group fairness. As a result, they can sometimes produce models that give blatantly \emph{unfair} predictions for specific individuals, even though they score well on group fairness metrics~\cite{dwork2012fairness, LiptonMC18}.

\paragraph{Counterfactual fairness.}
\emph{Counterfactual fairness}~\cite{kusner2017counterfactual}
requires that each prediction a classifier makes \emph{would have been
  the same} if the protected attribute had been different.
\emph{Counterfactual augmentation}~\cite{kaushik2019learning,
  kaushik2020explaining} focuses on counterfactual statements about
individuals in the original training data. Most approaches involve a
manual process of asking a human expert to construct a new training
example that is identical to an existing one, except for the protected
attribute. For example, in the NLP setting, classifiers often
associate the profession of nursing with
women~\cite{bolukbasi2016man}, due to the large volume of (biased)
statements in the training data that involve female nurses. For each
statement about a female nurse in the training data, we might ask the
expert to write an identical statement about a male nurse---and a
classifier trained on the augmented data will be less likely to
associate nursing with women as a result.

Counterfactual augmentation has proven extremely effective in models
for language~\cite{huang2020reducing, garg2019counterfactual,
  zmigrod2019counterfactual} and images~\cite{denton2019detecting},
where an explicit protected attribute is often not present. In these
domains, counterfactual augmentation is generally a manual process;
recent work provides support~\cite{madaan2020generate,
  wu2021polyjuice} but not complete automation.
Prediction sensitivity (defined in Section~\ref{sec:pred-sens}) can be
viewed as a way of measuring counterfactual fairness. Because our
evaluation is done in the simpler setting of a binary protected
attribute, we are able to use a relatively simple algorithm for
automatic counterfactual augmentation.

\paragraph{FlipTest.}
Our work is most closely related to FlipTest~\cite{black2020fliptest}, an approach for generating augmented datasets for auditing the counterfactual fairness of classifiers. FlipTest determines whether a model is sensitive to the protected status of an individual or subgroup using optimal transport mapping. It reveals salient patterns in a model's behavior by constructing approximate mappings using a Generative Adversarial Network (GAN). Using these mappings, flipsets and a transparency report are created, which are utilized to determine which individuals or subgroups are discriminated against, as well as identify those features which may be associated with it.

The major advantage of our approach is the ability to audit classifiers \emph{during deployment}. Unlike FlipTest, our approach does not require the explicit generation of datasets to measure counterfactual fairness; instead, it provides a measurement for each prediction the classifier makes, with no other inputs required. Our approach also has the novel ability to detect failures of counterfactual fairness resulting from previously-unseen data.

\paragraph{Interpretability in deep learning.}
The related field of \emph{interpretable AI} seeks to enable auditability by understanding \emph{why} a model made a specific prediction. The problem of interpretability has been studied extensively in the setting of image classification, where the goal is to understand which pixels of the image were most important in deciding its class. Numerous gradient-based approaches have been proposed~\cite{morch1995visualization, baehrens2010explain, simonyan2013deep, sundararajan2017axiomatic, smilkov2017smoothgrad, hooker2019benchmark}; our approach to prediction sensitivity draws on these ideas.



\section{Prediction Sensitivity}
\label{sec:pred-sens}

\begin{figure}
  \centering
  \includegraphics[width=.95\textwidth]{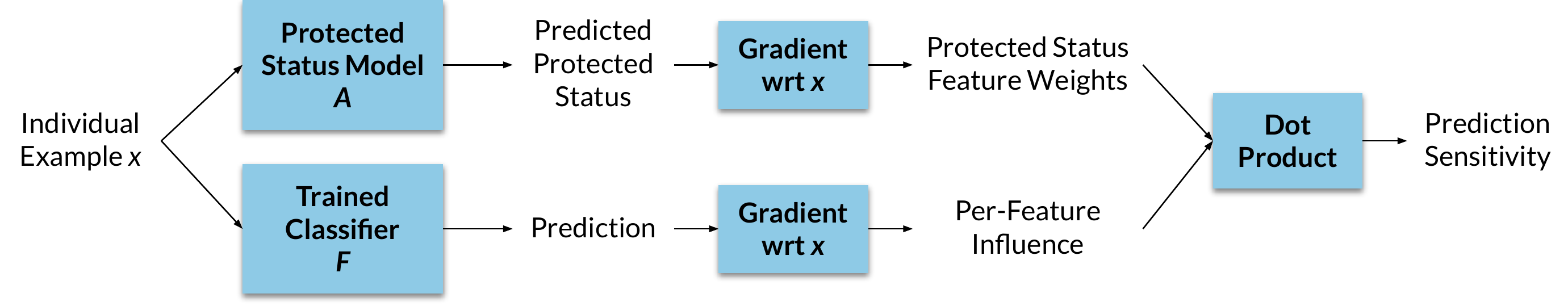}
  \caption{Overview of calculating prediction sensitivity. Prediction
    sensitivity is based on measurements of each feature's
    contribution to both protected status and the classifier's
    prediction.}
  \label{fig:overview}
\figspc
\end{figure}

This section describes prediction sensitivity, an easily calculated
metric for auditing the counterfactual fairness of deployed models.
For a single prediction, our goal is to answer the same question as
FlipTest~\cite{black2020fliptest}: \emph{had the individual been of a
  different protected status, would the model have treated them
  differently?}

We consider a classifier $\mathcal{F}$ and an individual input $\x$.
Let the \emph{counterfactual individual} $\x'$ represent the same
individual as $\x$, but with a different protected status. We would
like to know if $\mathcal{F}(\x) = \mathcal{F}(\x')$.
%
%
Prediction sensitivity answers this question by computing two
gradients with respect to the input $\x$, as depicted in
Figure~\ref{fig:overview}: the \emph{protected status feature weights}
(\S~\ref{sec:status_model}) and the \emph{per-feature influence}
(\S~\ref{sec:pred_sens_calc}). The final prediction sensitivity is the
dot product of the two components---a measure of how changes in the
input $\x$ would lead to changes in $\mathcal{F}$'s prediction,
weighted by each feature's influence on the individual's protected
status.
Both components are inspired by gradient-based approaches for
interpretability (especially {\sc
  SmoothGrad}~\cite{smilkov2017smoothgrad}).




\subsection{Protected Status Feature Weights}
\label{sec:status_model}


The \emph{protected status feature weights} measure \emph{how the
  features of $\x$ contribute to the $\x$'s protected status}. We
calculate the protected status feature weights by training a
\emph{protected status model} $\mathcal{A}$ to predict $\x$'s
protected status, and computing the weights using the gradient of
$\mathcal{A}(\x)$. If $\mathcal{A}$ captures the correlation between
input features and protected status, then these weights encode the
strength of each feature's value on protected status.


The protected status of an individual may be encoded in an explicit
protected attribute $a \in \x$, and $\mathcal{A}$ will learn that the
correlation between $a$ and the protected status.
When the protected attribute is {not} present at prediction time, or
other features are correlated with protected status, $\mathcal{A}$
will discover the appropriate correlations.
%
The specifics of the protected status model $\mathcal{A}$ depend on
the problem setting. In our experiments, we use a simple neural
network with a similar architecture to the classifier we are auditing.
The primary requirement is that $\mathcal{A}$ should capture
correlations between features in $\x$ and the protected status $s$.


\begin{definition}[Protected Status Feature Weights]
  The \emph{protected status feature weights} for an example $\x \in
  \mathbb{R}^k$ are a length-$k$ vector defined as follows:
  \[PSW(\x) = \textit{abs}(\nabla \mathcal{A}(\x))\]
  where $\textit{abs}$ denotes element-wise absolute value.
\end{definition}

\subsection{Per-Feature Influence \& Prediction Sensitivity}
\label{sec:pred_sens_calc}

The second component of prediction sensitivity is the
\emph{per-feature influence}, a vector that captures the influence of
each input feature on the classifier's prediction. The per-feature
influence describes \emph{how much a hypothetical change in each
  feature of $\x$ would affect $\mathcal{F}$'s prediction}. Prediction
sensitivity is defined as the dot product of this gradient with the
protected status feature weights for $\x$.

\begin{definition}[Prediction sensitivity]
  The \emph{prediction sensitivity} $PS(\x) \in \mathbb{R}$ for an
  example $\x$ is defined as:
  \begin{align*}
    PS(\x) &= PSW(\x) \cdot \textit{abs}(\nabla \mathcal{F}(\x))\\
           &= \textit{abs}(\nabla \mathcal{A}(\x)) \cdot \textit{abs}(\nabla \mathcal{F}(\x))
  \end{align*}
  where $\nabla \mathcal{A}(\x))$ is the gradient of
  $\mathcal{A}(\x))$ (with respect to $\x$) and $\textit{abs}$ denotes
  element-wise absolute value.
\end{definition}


\subsection{Prediction Sensitivity as a Measure of Counterfactual Fairness}

Prediction sensitivity is designed to answer the question: \emph{had
  the individual been of a different protected status, would the model
  have treated them differently?} 
Prediction sensitivity is thus likely to be a close approximation of a
measurement of counterfactual fairness in most settings. Formally,
counterfactual fairness is defined as follows:

\begin{definition}[Counterfactual fairness~\cite{kusner2017counterfactual}]
  A predictor $\hat{Y}$ of $Y$ is \emph{counterfactually fair} given the
  sensitive attribute $A = a$ and any observed variables $X$ if:
  \[\Pr[ \hat{Y}_{A \leftarrow a} = y \;|\; X = x, A = a] = 
    \Pr[ \hat{Y}_{A \leftarrow a'} = y \;|\; X = x, A = a]\]
  for all $y$ and $a' \neq a$ (i.e. $a'$ denotes a different protected
  status than $a$).
\end{definition}

This definition requires that the distribution of $\hat{Y}$ does not
change when $A$ changes, as long as things which are not causally
dependent on $A$ are held constant. Demonstrating counterfactual
fairness thus requires knowledge about the causal relationships
between variables (in particular, about the causal relationships
between $A$ and other attributes).

In our approach, the protected status model $\mathcal{A}$ encodes
correlations between the features of $\x$ and the protected status
($A$ in the above definition). Our $\mathcal{A}$ model can be seen as
an overapproximation of the required causality knowledge, in the sense
that it may learn correlations that are not causal. Prediction
sensitivity therefore may produce false positives (i.e. it may be
high, even for a prediction that satisfies counterfactual fairness),
but as long as the $\mathcal{A}$ model correctly captures correlations
with the protected status, it will not produce false negatives.

A second important condition for the correspondence between prediction
sensitivity and counterfactual fairness is that the gradient should
provide good information about the influence of individual features on
the prediction (for both the $\mathcal{A}$ and $\mathcal{F}$ models).
When the gradient is smooth, then prediction sensitivity should be
an effective way to measure counterfactual fairness for individual
predictions.

\section{Using Prediction Sensitivity to Audit Deployed Models}
\label{sec:using-pred-sens}

\begin{figure}
\centering
\includegraphics[width=.98\textwidth]{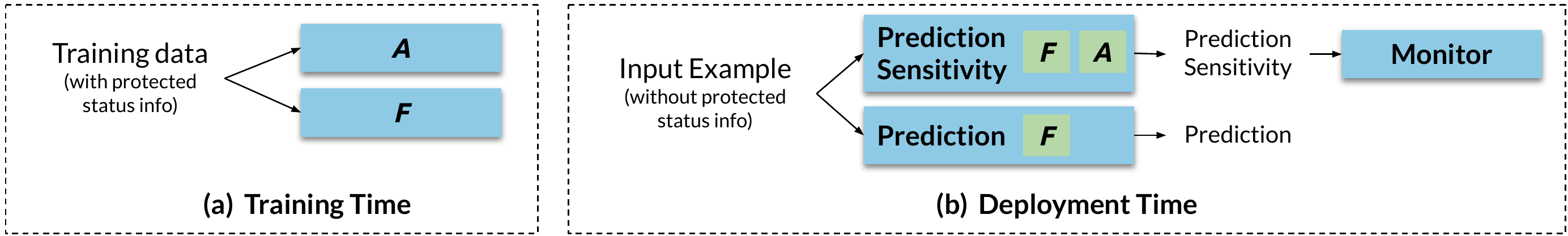}

\caption{Continual audit of counterfactual fairness on a deployed
  classifier using prediction sensitivity.}
\label{fig:deploy}
\figspc
\end{figure}

Prediction sensitivity can be calculated at training time, using a
test set, to measure fairness properties of the trained classifier.
However, the real strength of prediction sensitivity is that it can
also be used \emph{at prediction time}---calculating prediction
sensitivity requires neither the true label nor the protected
attribute.
Prediction sensitivity can thus be deployed \emph{alongside}
production classifiers, to continually measure fairness properties of
the predictions those classifiers make over time, as shown in
Figure~\ref{fig:deploy}.

\paragraph{Training time.}
At training time, we train both the classifier $\mathcal{F}$ and the
protected status model $\mathcal{A}$, as shown in
Figure~\ref{fig:deploy}(a). Training $\mathcal{A}$ requires
information about protected status to be present in the training data.
The classifier $\mathcal{F}$ can be trained to satisfy fairness
conditions using training-time techniques
(e.g.~\cite{calders2009building, woodworth2017learning,
  zafar2015fairness, zafar2017fairness, agarwal2018reductions,
  hardt2016equality}), and can be audited before it is deployed using
group fairness metrics and approaches like
FlipTest~\cite{black2020fliptest}).

We can also use prediction sensitivity to audit the trained classifier
for counterfactual fairness at training time. We compute prediction
sensitivity for examples in the test set and ensure that (1) both the
mean and variance of the prediction sensitivity are low, and (2) there
are no outliers representing disadvantaged individuals. The mean
prediction sensitivity for the classifier on the test set can be saved
as a \emph{baseline} for use during deployment.

\paragraph{Deployment time.}
After the classifier is deployed, prediction sensitivity can be
calculated for each prediction and compared against the baseline by a
monitor deployed with the system, as shown in
Figure~\ref{fig:deploy}(b). If a particular prediction results in
prediction sensitivity much higher than the baseline, then it is
likely that counterfactual fairness has been violated. In this case,
the monitor can raise an alarm.

\paragraph{Reacting to high prediction sensitivity.}
When the classifier fails to satisfy counterfactual fairness,
prediction sensitivity can help guide action to correct the situation.
\begin{enumerate}[topsep=0pt, leftmargin=16pt]
\item Most importantly, the prediction should not be used to make
  decisions, since it may cause harm. The system should fall back on
  manual intervention by an expert, or a backup system (e.g. a default
  decision).
\item Properties of the individual whose prediction caused the failure
  of counterfactual fairness may immediately suggest an approach for
  improving the classifier $\mathcal{F}$. Such failures may be due to
  a lack of similar individuals in the training data; here, the
  classifier can be improved by re-training on additional data.
\item The feature-wise prediction sensitivity may indicate that the
  classifier has focused on specific features in making discriminatory
  predictions. Particular attention should be paid to these features
  when improving the training data.
\end{enumerate}

\paragraph{Auditing by third parties.}
Prediction sensitivity can only be calculated with whitebox access to
the classifier $\mathcal{F}$ and protected status model $\mathcal{A}$,
because of the gradient calculations involved. This limits the use of
prediction sensitivity to the party who trained and deployed the
model; it is not feasible for third parties to audit a deployed system
using prediction sensitivity without access to the classifiers
involved.

This limitation means that prediction sensitivity is not a useful tool
for helping journalists or other third parties to discover
discriminatory behavior in existing systems. However, corporations
face increasing public pressure to ensure that their systems do not
discriminate; this pressure provides a significant incentive for
companies to adopt approaches like prediction sensitivity to improve
their models and provide evidence of fairness to their users.




\section{Evaluation on Synthetic Data via Causal Modeling}
\label{sec:synthetic-eval}


This section describes an empirical evaluation of prediction
sensitivity as a measure for counterfactual fairness using synthetic
data. Using synthetic data allows us to precisely model our desired
fairness definition (including causal information), and measure
the effectiveness of our approach.

\subsection{Evaluation Approach}

Prediction sensitivity is intended to answer the question, \emph{had
  the individual been of a different protected status, would the model
  have treated them differently?} Given a dataset of otherwise
identical individuals with different protected statuses, we could ask
our model to make a prediction for each one and compare them;
differing predictions for corresponding individuals represent failures
of counterfactual fairness.
However, actually \emph{constructing} this dataset is typically very
challenging~\cite{dwork2012fairness, black2020fliptest}. By employing
synthetic data, we can use an alternative approach that avoids this
challenge but accomplishes the same goal.

Our approach is to train two classifiers and then compare their
predictions to detect failures of counterfactual fairness. One is the
``original'' classifier $\mathcal{F}$, which may not satisfy
counterfactual fairness. The second model is the \emph{unbiased
  classifier} $\hat{\mathcal{F}}$, trained on modified data to ensure
counterfactual fairness.
We evaluate the counterfactual fairness of $\mathcal{F}$ by comparing
its predictions on the test set against those of $\hat{\mathcal{F}}$.
When the two models make \emph{different} predictions, this indicates
a possible failure of counterfactual fairness. We call the set of
individuals in the test set for whom the two models make matching
predictions the \emph{match set}.
We evaluate prediction sensitivity as a measure of counterfactual
fairness by checking whether prediction sensitivity can effectively
distinguish members of the match set from non-members.

\begin{figure}
\centering
{\small
  \begin{tabular}{c c c}
    \includegraphics[width=.45\textwidth]{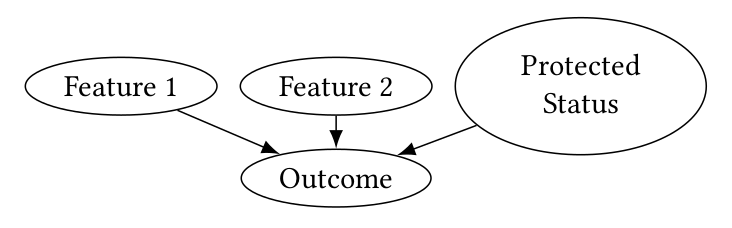}
&
\phantom{hi}
&
    \includegraphics[width=.45\textwidth]{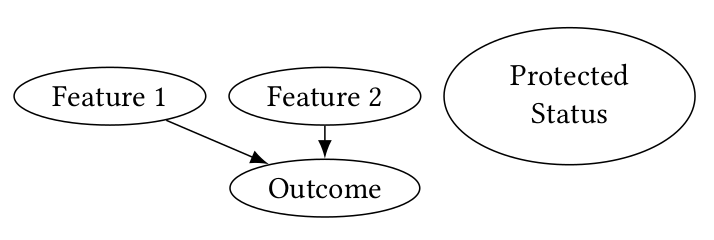}
\\[10pt]
\textbf{(a)}
&&
\textbf{(b)}\\
\end{tabular}
}
\caption{Causal graphs for synthetic data. \textbf{(a)} shows a causal
  graph for ``biased'' synthetic data, in which a causal relationship
  exists between protected status and outcome. \textbf{(b)} shows a
  modified causal graph that removes this relationship. Data generated
  according to model (b) can be used to train classifiers that satisfy
  counterfactual fairness.}
\label{fig:causal_models}
\figspc
\end{figure}

\paragraph{Modeling counterfactual fairness in synthetic data.}
%
We generate two synthetic datasets
according to the causal models shown in
Figure~\ref{fig:causal_models}. The causal model in
Figure~\ref{fig:causal_models}(a) includes a causal relationship
between protected status and outcome, and is thus likely to result in
biased classifiers. In our evaluation, we train the classifier
$\mathcal{F}$ using data generated according to this model---so we
expect that $\mathcal{F}$ will not always satisfy counterfactual
fairness. The causal model in Figure~\ref{fig:causal_models}(b) lacks
this causal relationship. We train the \emph{unbiased classifier}
$\hat{\mathcal{F}}$ using data generated according to this model, so
that $\hat{\mathcal{F}}$ satisfies counterfactual fairness.



In our evaluation, following Lipton et al.~\cite{LiptonMC18}, we
generate a synthetic dataset with two features and binary labels.
Augmenting this dataset with a protected attribute (gender) chosen at
random will produce a dataset consistent with the causal model in
Figure~\ref{fig:causal_models}(b).
%
Models trained on this dataset score well on group fairness metrics
like statistical parity and disparate impact, since gender is
completely independent of the label and other features. We use this
dataset to train the model $\hat{\mathcal{F}}$.
We introduce bias in our synthetic dataset by generating the protected
attribute based on the label:
\[ \Pr(z = \text{woman}) = \left\{
	\begin{array}{ll}
		.25  & \mbox{if } y = \text{positive} \\
		.75 & \mbox{if } y = \text{negative} 
	\end{array}
\right. \]
This model clearly introduces bias against women---men are much more
likely to be members of the positive class, while women are much more
likely to be members of the negative class. We use this dataset to
train the model $\mathcal{F}$.

\paragraph{Enumerating match sets.}
The next step is to find individuals for whom $\mathcal{F}$ and
$\hat{\mathcal{F}}$ make different predictions. Since
$\hat{\mathcal{F}}$ is defined to satisfy counterfactual fairness,
such a mismatch likely indicates a \emph{failure} of counterfactual
fairness for $\mathcal{F}$.
We call the set of individuals for whom both models make the same
predictions a \emph{match set}. The classifier $\mathcal{F}$ likely
satisfies counterfactual fairness for members of the match set, and
likely violates it for non-members. For counterfactually fair
classifiers, the match set should include the entire test set.

\paragraph{Evaluating prediction sensitivity.}
The final step of our evaluation is to demonstrate that prediction
sensitivity is effective for distinguishing members of the match set
from non-members.
We calculate the prediction sensitivity for $\mathcal{F}$'s
predictions on the test set, and use it to build a classifier
$\mathcal{D}$ to distinguish match set members from non-members.
$\mathcal{D}$ is defined as follows (where $p$ is the prediction
sensitivity associated with an example, and $\theta$ is a threshold):
\[ \mathcal{D}(p, \theta) = \left\{
	\begin{array}{ll}
		\textbf{member}  & \mbox{if } p \leq \theta \\
		\textbf{non-member} & \mbox{if } p > \theta
	\end{array}
\right. \]
Here, the threshold $\theta$ allows trading off between false
negatives and false positives.

\paragraph{Comparison to FlipTest evaluation.}
FlipTest~\cite{black2020fliptest} calculates a similar set of
individuals---called a \emph{flipset}---by directly generating
datasets of counterfactuals. For each individual $\x$ in the test set,
FlipTest uses a generative adversarial network (GAN) to generate an
in-distribution individual $\x'$ with the opposite protected status.
Then, FlipTest asks the classifier under evaluation ($\mathcal{F}$) to
make predictions on each individual and their corresponding
counterfactual. Matching predictions are considered counterfactually
fair, since the model treats both individuals the same; non-matching
predictions are considered failures of counterfactual fairness.
FlipTest calls the set of non-matching predictions a
\emph{flipset}---the set of individuals for whom the prediction flips
when protected status changes. Our approach has a similar goal---the
flipset serves the same purpose as our match set, and inspires its
name---but our approach does not require the construction of a GAN or
the enumeration of counterfactual individuals.

\subsection{Experiment Setup}

We used Scikit-learn to generate synthetic datasets consistent with
the causal models shown in Figure~\ref{fig:causal_models}. We trained
the classifier $\mathcal{F}$ using data consistent with the model in
Figure~\ref{fig:causal_models}(a), and the counterfactually fair
classifier $\hat{\mathcal{F}}$ using data consistent with the model in
Figure~\ref{fig:causal_models}(b). We trained $\mathcal{A}$,
$\mathcal{F}$, and $\hat{\mathcal{F}}$, then computed prediction
sensitivities and constructed the distinguisher $\mathcal{D}$
consistent with the process described in the last section.
We constructed receiver operating characteristic (ROC) curves and
calculated area under the curve (AUC) values for the distinguisher by
varying $\theta$.
Due to nondeterminism in the training process, we performed 30 trials
of the experiment.

\subsection{Results}

\begin{figure}
\centering
\begin{tabular}{c c}
  \includegraphics[width=.40\textwidth]{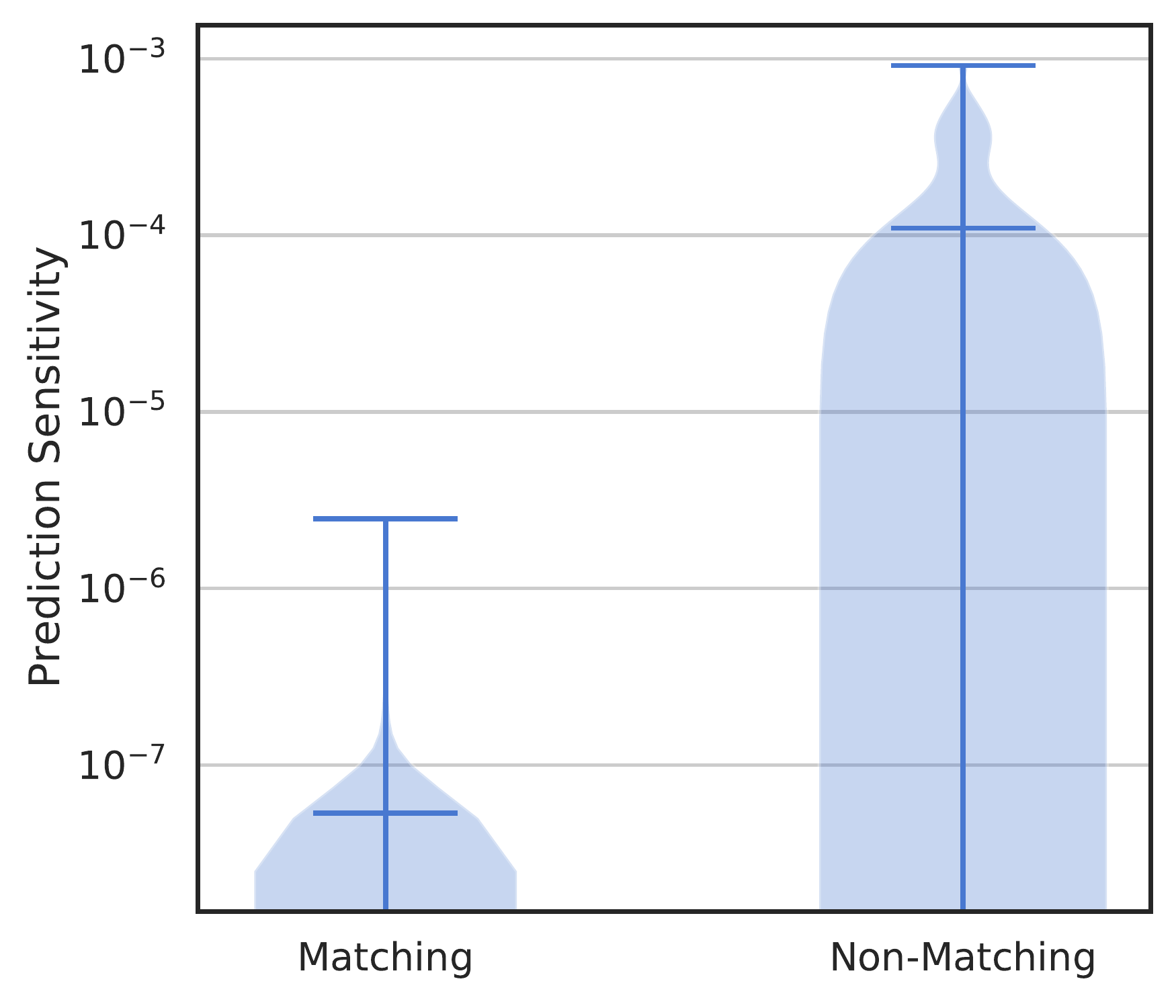} 
  & \includegraphics[width=.45\textwidth]{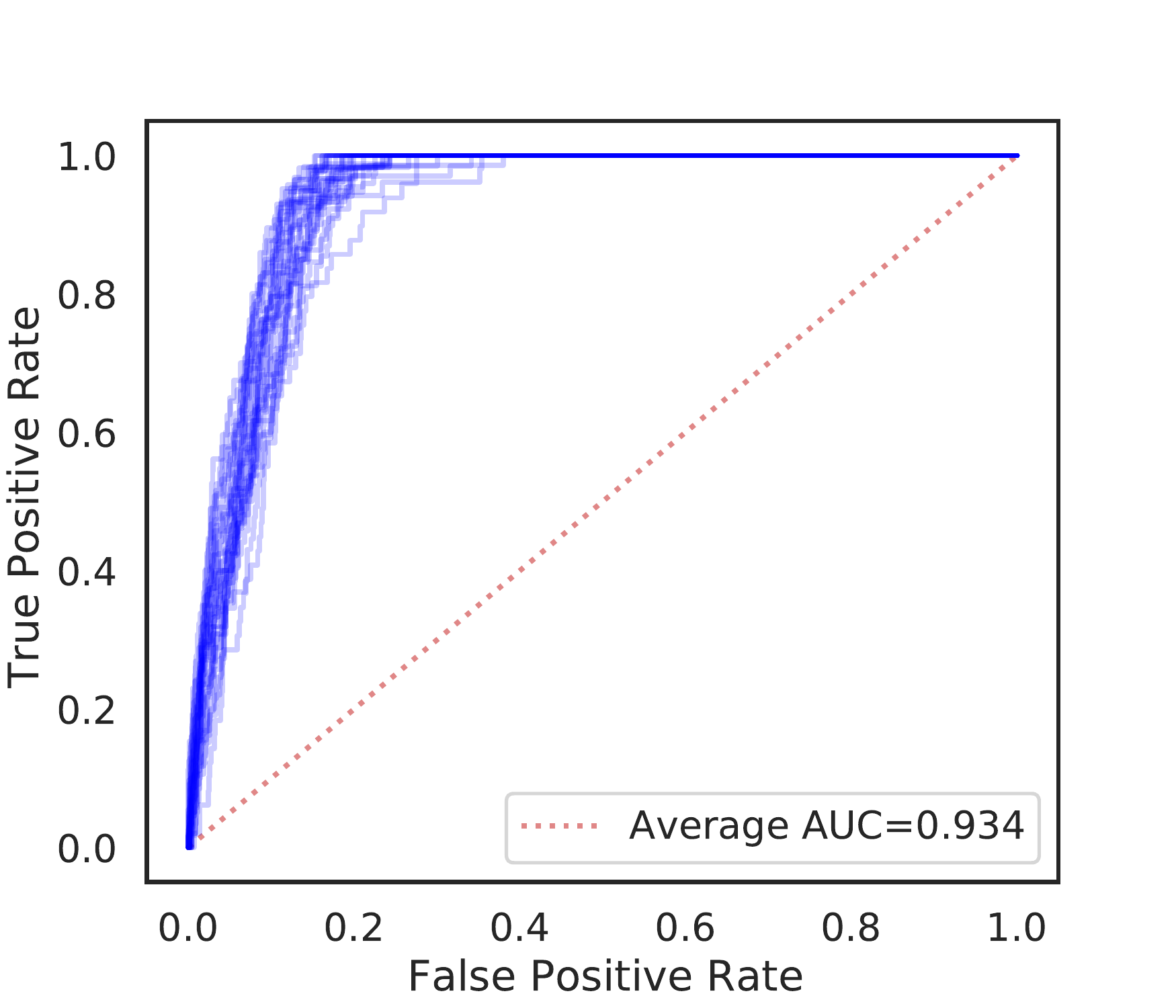} \\
  \textbf{(a)} & \textbf{(b)}
\end{tabular}

\caption{Using prediction sensitivity to audit models trained on
  synthetic data. \textbf{(a)} shows that prediction sensitivity is
  low for members of the match set, but high for non-members (note the
  logarithmic scale in the vertical axis). \textbf{(b)} shows that a
  distinguisher based on prediction sensitivity is effective at
      detecting failures of counterfactual fairness.}
\label{fig:synthetic_results}
\figspc
\end{figure}

The results of our experiment on synthetic data appear in
Figure~\ref{fig:synthetic_results}.
Figure~\ref{fig:synthetic_results}(a) shows that average prediction
sensitivity is much higher for non-members of the match set than it is
for members, suggesting that prediction sensitivity is able to
distinguish individuals for whom the classifier $\mathcal{F}$ fails to
satisfy counterfactual fairness.

Figure~\ref{fig:synthetic_results}(b) shows a receiver operating
characteristic (ROC) curve and its area under the curve (AUC) value
for the distinguisher $\mathcal{D}$ built using prediction
sensitivity. We plot one curve per trial we performed (30 curves in
total) and give the average AUC over all trials. The results show that
prediction sensitivity is capable of distinguishing members of the
match set from non-members, even for low false positive rates.

\section{Evaluation on Real Data via Counterfactual Augmentation}

In Section~\ref{sec:synthetic-eval}, we were able to generate
synthetic data consistent with causal models we designed ourselves,
enabling the comparison of the two models ($\mathcal{F}$ and
$\hat{\mathcal{F}}$) to build match sets. A similar process is
impossible for real data, since the underlying causal model is
unknown.

For our evaluation on real datasets, we instead approximate the same
procedure using \emph{counterfactual augmentation}. {Counterfactual
  augmentation}~\cite{kaushik2019learning, kaushik2020explaining}
attempts to ensure counterfactual fairness in trained classifiers by
adding new training examples. For each individual $\x$ in the training
data, we add a corresponding individual $\x'$ with a \emph{different
  protected status} but the \emph{same outcome}. Classifiers trained
on the augmented data will be more likely to satisfy counterfactual
fairness, because each training example has a corresponding
counterfactual also present during training. Previous work has shown
this approach to be effective in training classifiers that also score
well on group fairness metrics~\cite{kaushik2019learning,
  kaushik2020explaining}.

\subsection{Evaluation Approach}

Our evaluation on real data follows the same process as our synthetic
data evaluation described in Section~\ref{sec:synthetic-eval}, but
using counterfactual augmentation to construct training data for the
counterfactually fair model $\hat{\mathcal{F}}$.

  

  



\paragraph{Constructing augmented datasets.}
In complex domains like NLP, counterfactual augmentation is a manual
process. In our simpler setting of binary classification with an
explicit, binary protected attribute, it is possible to automate this
process. In particular, for each example in the original dataset, we
can construct a new example with the opposite value for the
protected attribute.
%
For each training example $\x$, we construct a new one $\x'$ in which
the protected attribute is replaced by its negation. The label remains
the same. The resulting dataset has exactly twice the original number
of examples.
%
This approach is simple and effective, but it requires the protected
attribute to be discrete (ideally, binary) and be explicitly included
in the data (in e.g. NLP data, it often is not).

\paragraph{Evaluating prediction sensitivity.}
Having constructed an augmented dataset, we can evaluate prediction
sensitivity in the same way as in Section~\ref{sec:synthetic-eval}, by
constructing a match set and a distinguisher.

\paragraph{Limitations of our implementation of counterfactual augmentation.}
The goal of counterfactual augmentation is to approximate data drawn
from a similar distribution to the original data, but \emph{without}
the causal relationship between protected status and outcome.
However, our approach modifies only the protected attribute, and
leaves other (possibly correlated) features alone. If other features
are correlated with the protected attribute, then the correlations
between those features and the individual's \emph{original} protected
status may still exist.

Consider the Adult dataset~\cite{adult} (used in our empirical
evaluation), which includes job category and gender (a protected
attribute) as features, and income as the label. It is likely that job
category is correlated with gender, but our approach for
counterfactual augmentation ignores this correlation.
As a result, the classifier $\hat{\mathcal{F}}$ may \emph{fail} to
satisfy counterfactual fairness in some cases, even though it is
trained on counterfactually augmented data, because it may learn to
discriminate based on features correlated with the protected attribute
(rather than the protected attribute itself).
%


This is a limitation of our evaluation approach---not of prediction
sensitivity itself. As described in Section~\ref{sec:pred-sens},
prediction sensitivity does capture correlations between all features
and protected status, via the protected status model.

\paragraph{Comparison to FlipTest evaluation.}
As described in Section~\ref{sec:synthetic-eval}, the match sets we
generate in our evaluation are similar to FlipTest's flipsets.
However, FlipTest uses a generative adversarial network (GAN) to
generate the augmented data to construct flipsets, while we use a
simpler form of counterfactual augmentation. The GAN-based approach
can better capture correlations between the protected attribute and
other features, but it is more complex and may also fail to generate
out-of-distribution counterfactuals that would have exposed failures
in the classifier being tested.

\subsection{Experiment Setup}
\label{sec:experiment-setup}

We follow the same experimental setup as in
Section~\ref{sec:synthetic-eval}, training $\mathcal{F}$ on the
original data and $\hat{\mathcal{F}}$ on the augmented data.
We calculated prediction sensitivity for each of $\mathcal{F}$'s
predictions on the test set as described in
Section~\ref{sec:pred-sens}, training the protected status model
$\mathcal{A}$ using the original training data used for $\mathcal{F}$.
We defined the distinguisher $\mathcal{D}$ based on these prediction
sensitivity values.
As in Section~\ref{sec:synthetic-eval}, we performed 30 trials.

\paragraph{Datasets.}
Our evaluation considers two commonly-used datasets in the AI fairness
literature: the Adult~\cite{adult} and COMPAS~\cite{compas}
datasets. Statistics of the two datasets are shown in
Table~\ref{tbl:datasets}. Both involve classification tasks: the
Adult dataset's task is predicting whether an individual's income is
greater than or less than \$50,000, and the COMPAS dataset's task is
predicting recidivism risk. Both datasets are known to contain
embedded bias: models trained on the Adult dataset tend to predict
that white males have the highest chance of having high income, and
models trained on the COMPAS dataset tend to predict that Black
suspects are most likely to re-offend.
We augment both datasets with counterfactual training examples as
described earlier. This process produces datasets that are exactly
twice as large as the originals. For each male instance in the Adult
dataset, for example, we add a female instance that is identical
(including the label) except for the protected attribute.

\begin{table}
  \centering
  \begin{tabular}{|c||c c c||c c c c|}
    \hline
    \textbf{Dataset}
    & \multicolumn{3}{c||}{\textbf{Dataset Size}}
    & \multicolumn{4}{c|}{\textbf{Model Properties}}
    \\
    & Features & Train & Test & Layers & Layer Width & Activation & Accuracy\\
    \hline
    \hline
    Adult~\cite{adult}
    & 95
    & 24752 
    & 6188  
    & 3
    & 32
    & ReLU
    & 84.6\%
    \\
    COMPAS~\cite{compas}
    & 399
    & 4937  
    & 1235  
    & 3
    & 256
    & ReLU
    & 67.9\%
    \\
    \hline
  \end{tabular}
  \caption{Information about the datasets and models used in our experiments.}
  \label{tbl:datasets}
  \figspc
\end{table}

\paragraph{Model architectures.}
As shown in Table~\ref{tbl:datasets}, we use a linear network
architecture with a single hidden layer and the ReLU activation
function. The sigmoid activation function is applied on the output
layer. The loss function is the binary cross entropy loss and Adam is
chosen as the optimizer with a learning rate of 0.001. We find that a
wider model (256 neurons) works better for the COMPAS dataset, while a
narrower one (32 neurons) suffices for the Adult dataset. We train
each classifier for 40 epochs, and achieve roughly state-of-the-art
accuracy for both datasets.

\paragraph{Test sets.}
When evaluating the fairness of a classifier, should we also augment
the test set via counterfactual augmentation? Wick et
al.~\cite{NEURIPS2019_373e4c5d} argue in favor of an \emph{unbiased}
test set (i.e. ``fair'' data) to evaluate fairness. In particular, a
fair classifier may produce lower test accuracy on biased data,
leading to the (false) conclusion that improvements to fairness have
``hurt'' accuracy.
We ran our experiments twice: once using test sets containing original
data only (20\% of the original dataset), and once using
counterfactually augmented test sets. We report results for the
augmented test sets, since we found essentially no differences between
the two settings (more in Section~\ref{sec:exp2}).



\begin{figure}
  \centering
  \begin{tabular}{c c c}
    \hspace*{-1em}\includegraphics[width=.30\textwidth]{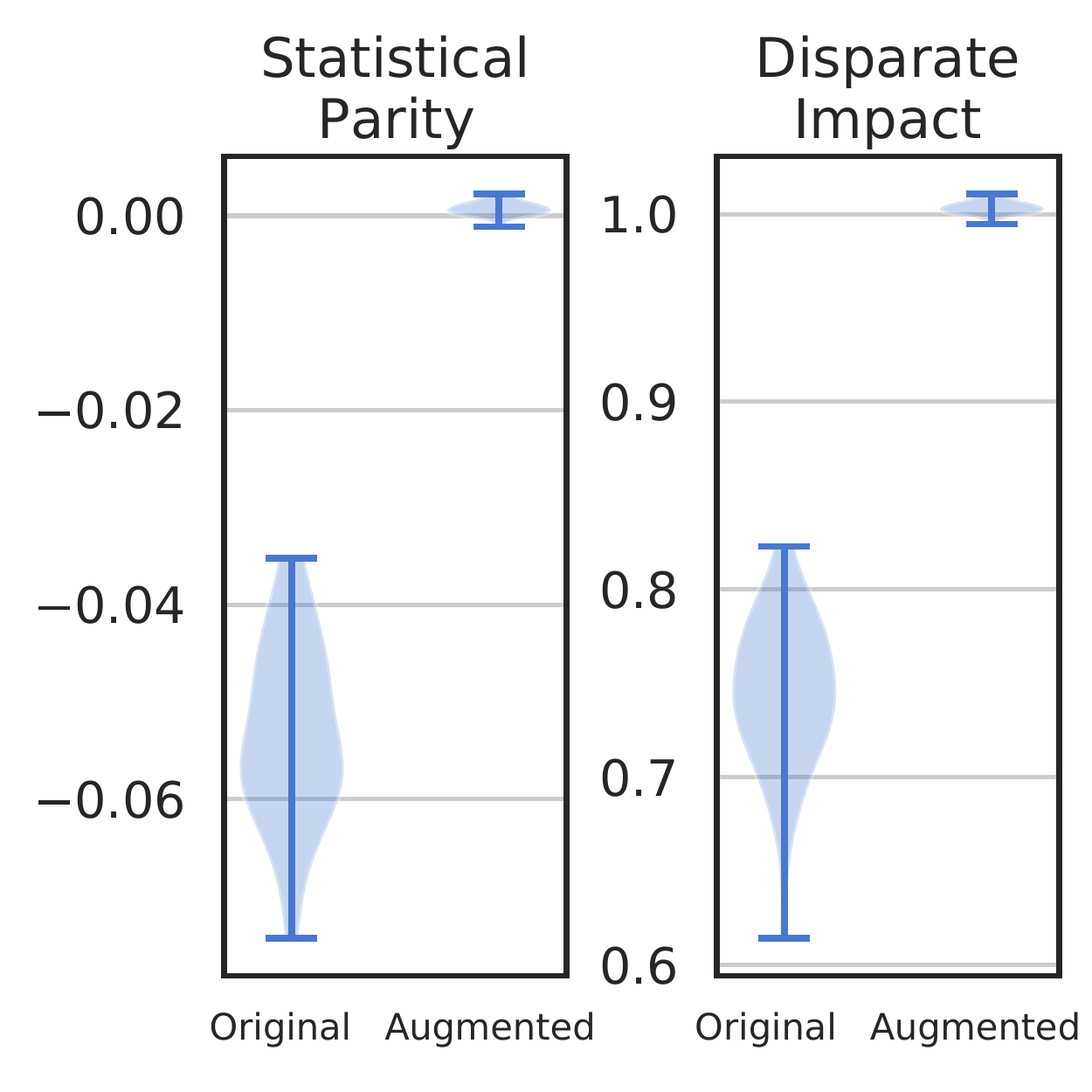}
    \hspace*{2em}
    & \hspace*{-1em}\includegraphics[width=.30\textwidth]{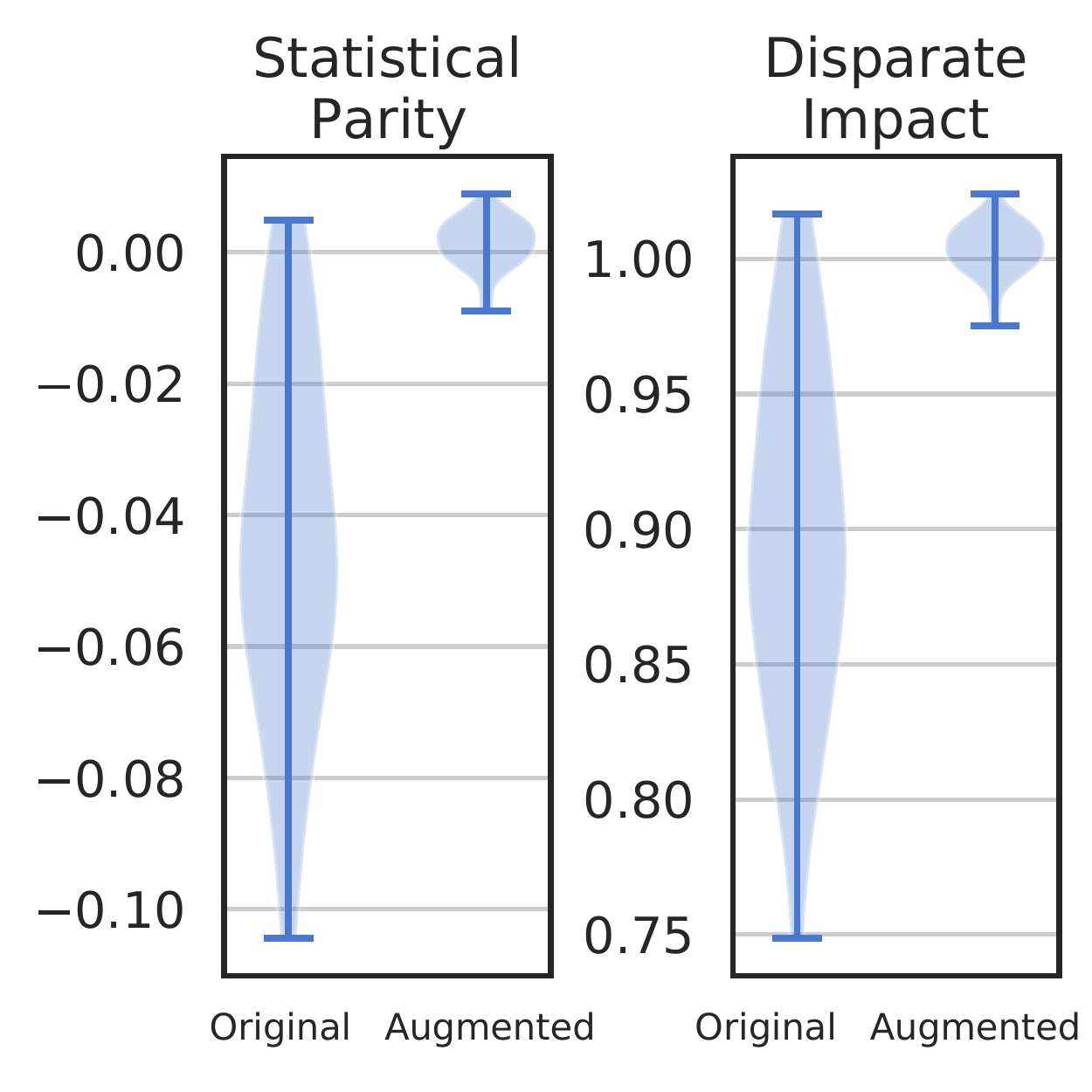}
      \hspace*{2em}
    & \hspace*{-1em}\includegraphics[width=.30\textwidth]{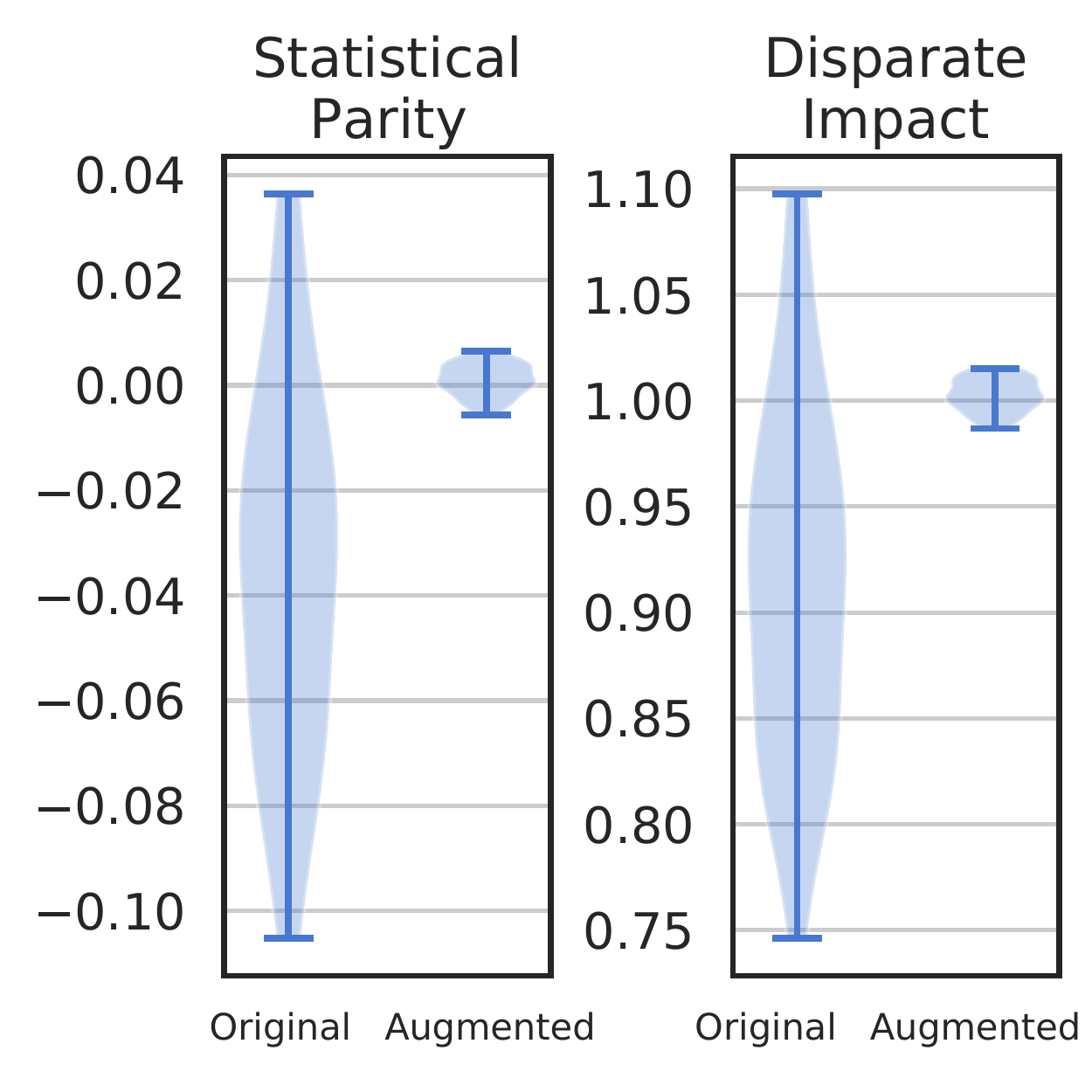}
      \hspace*{2em}\\
    \textbf{Adult (sex)}
    & \textbf{COMPAS (sex)}
    & \textbf{COMPAS (race)} \\
  \end{tabular}
  \caption{Distributions of group fairness metrics for real datasets,
    based on 30 trials for each configuration. As expected, models
    trained with counterfactually augmented data score much better on
    group fairness metrics.}
  \label{fig:group_fairness}
  \figspc
\end{figure}

\subsection{Results}
\label{sec:exp2}

\begin{table}
  \centering
  \begin{tabular}{|l c |ll |ll|}
    \hline
    \textbf{Dataset \&} 
    & \textbf{Training} 
    & \multicolumn{2}{c|}{\textbf{Classifier Accuracy}} 
    & \multicolumn{2}{c|}{\textbf{Prediction Sens. AUC}} \\
    \textbf{Prot. Attr.} & \textbf{Epochs} 
    & Original & Augmented
      & Original & Augmented \\
    \hline
    \hline
    \multirow{2}{*}{Adult~\cite{adult} (sex)}
    & 10
    & 84.0\% $\pm$ 1.65
    & 84.3\% $\pm$ 1.53
    & 57.4\% $\pm$ 0.17
    & 59.3\% $\pm$ 0.15
    \\
    & 40
    & 84.4\% $\pm$ 1.52
    & 85.1\% $\pm$ 0.84
    & 69.8\% $\pm$ 0.15
    & 69.9\% $\pm$ 0.17
    \\
    \hline
    \multirow{2}{*}{Adult~\cite{adult} (race)}
    & 10
    & 83.4\% $\pm$ 1.74
    & 84.1\% $\pm$ 1.82
    & 57.5\% $\pm$ 0.18
    & 60.4\% $\pm$ 0.17
    \\
    & 40
    & 84.6\% $\pm$ 1.55
    & 84.0\% $\pm$ 2.13
    & 66.7\% $\pm$ 0.19
    & 66.3\% $\pm$ 0.19
    \\
    \hline
    \multirow{2}{*}{COMPAS~\cite{compas} (sex)}
    & 10
    & 67.7\% $\pm$ 0.60
    & 66.8\% $\pm$ 0.74
    & 77.7\% $\pm$ 0.07
    & 79.4\% $\pm$ 0.06
    \\
    & 40
    & 66.7\% $\pm$ 0.54
    & 66.0\% $\pm$ 0.79
    & 76.4\% $\pm$ 0.05
    & 77.5\% $\pm$ 0.04
    \\
    \hline
    \multirow{2}{*}{COMPAS~\cite{compas} (race)}
    & 10
    & 67.9\% $\pm$ 0.40
    & 66.9\% $\pm$ 0.42
    & 79.2\% $\pm$ 0.07
    & 78.0\% $\pm$ 0.07
    \\
    & 40
    & 66.6\% $\pm$ 0.47
    & 65.9\% $\pm$ 0.47
    & 74.8\% $\pm$ 0.05
    & 75.1\% $\pm$ 0.05
    \\
    \hline
  \end{tabular}
  \caption{Comparison of average area under the curve (AUC) for
    original and augmented test sets. Results averaged over 30
    trials.}
  \label{tbl:full_numbers}
  \figspc
\end{table}

\begin{figure}
  \newcommand{\rmhspc}{\hspace*{-10pt}}
  \centering
  \begin{tabular}{c c c c}
    \includegraphics[width=.24\textwidth]{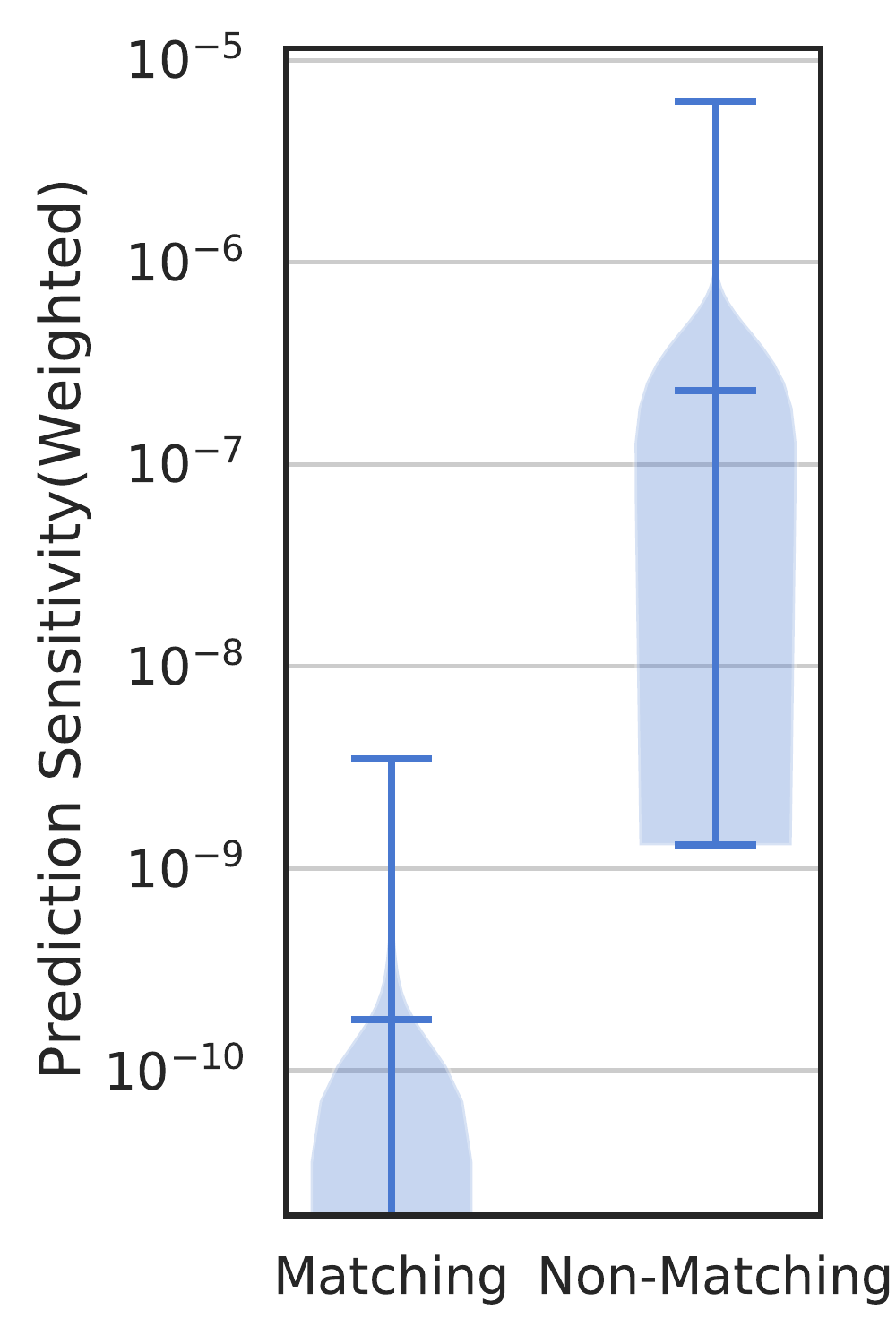}
    \rmhspc
    & \includegraphics[width=.24\textwidth]{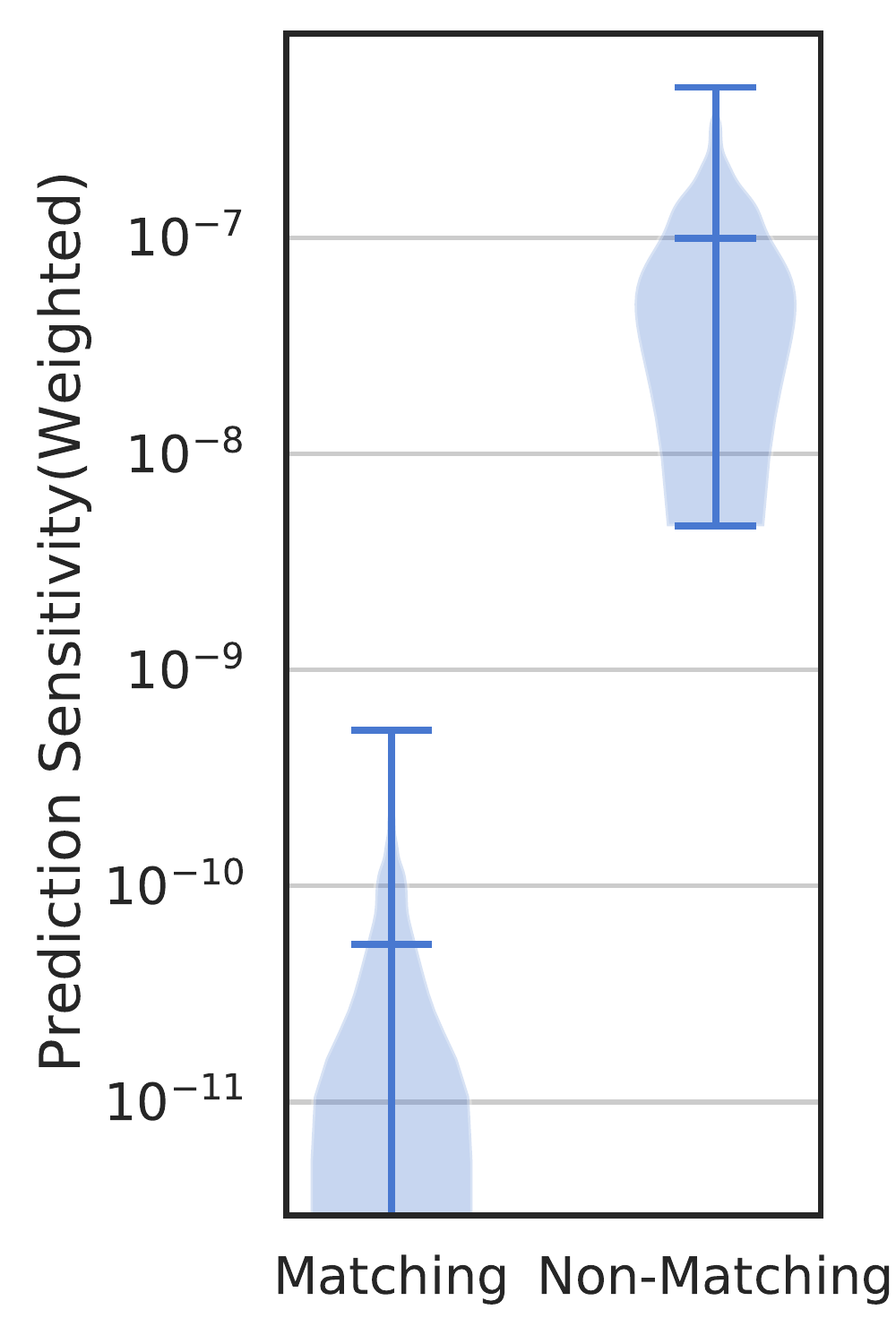}
    \rmhspc
    & \includegraphics[width=.24\textwidth]{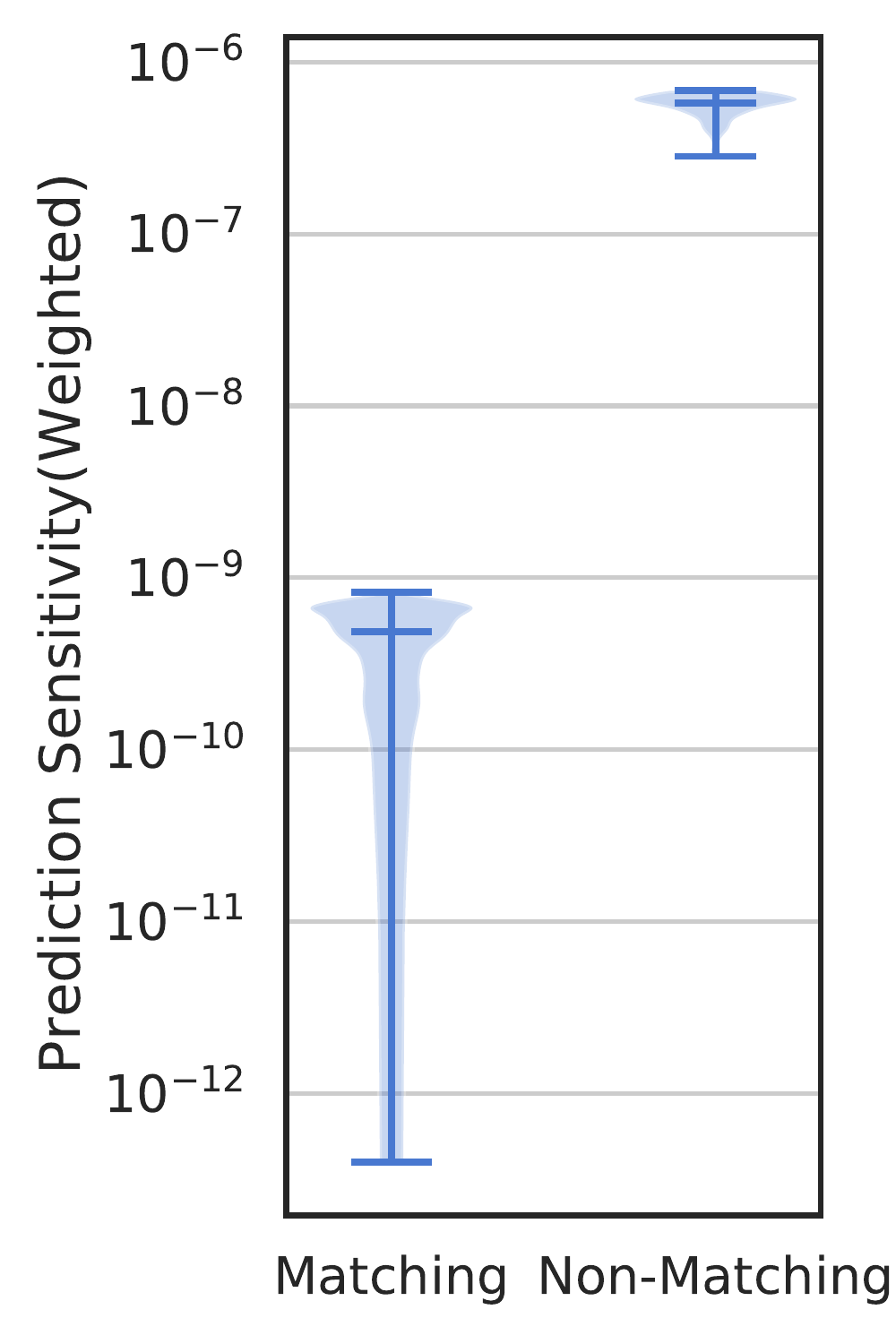}
    \rmhspc
    & \includegraphics[width=.24\textwidth]{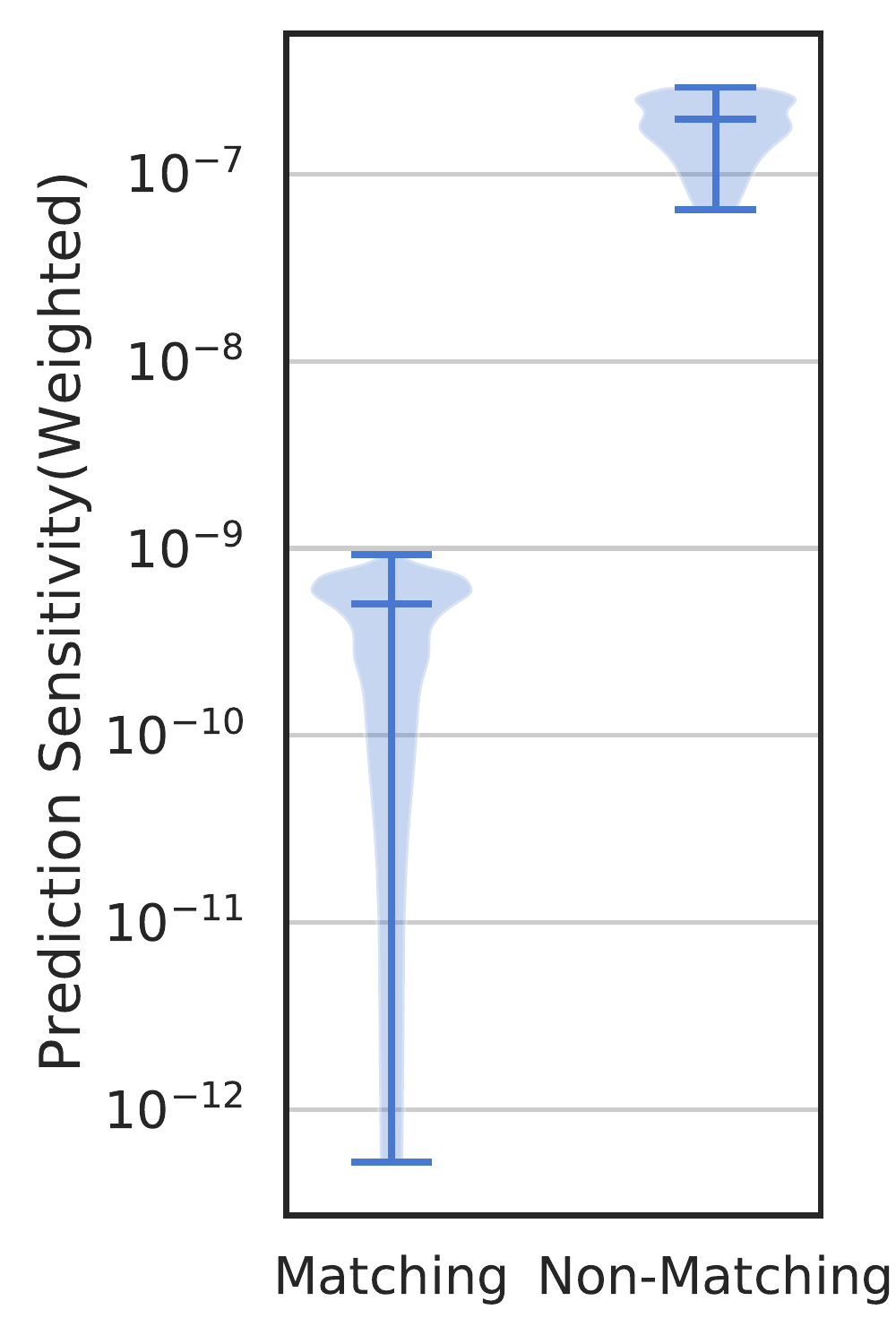}
    \\
    \textbf{Adult (sex)}
    & \textbf{Adult (race)}
    & \textbf{COMPAS (sex)}
    & \textbf{COMPAS (race)} \\
  \end{tabular}
  \caption{Distributions of prediction sensitivities for real
    datasets, based on 30 trials for each configuration. Note the
    logarithmic scale of the vertical axis. In all cases, average
    prediction sensitivity is much higher for non-members of the match
    set than it is for members.}
  \label{fig:mean_sensitivity}
  \figspc
\end{figure}

\begin{figure}
  \centering
  \begin{tabular}{c c}
    \includegraphics[width=.5\textwidth]{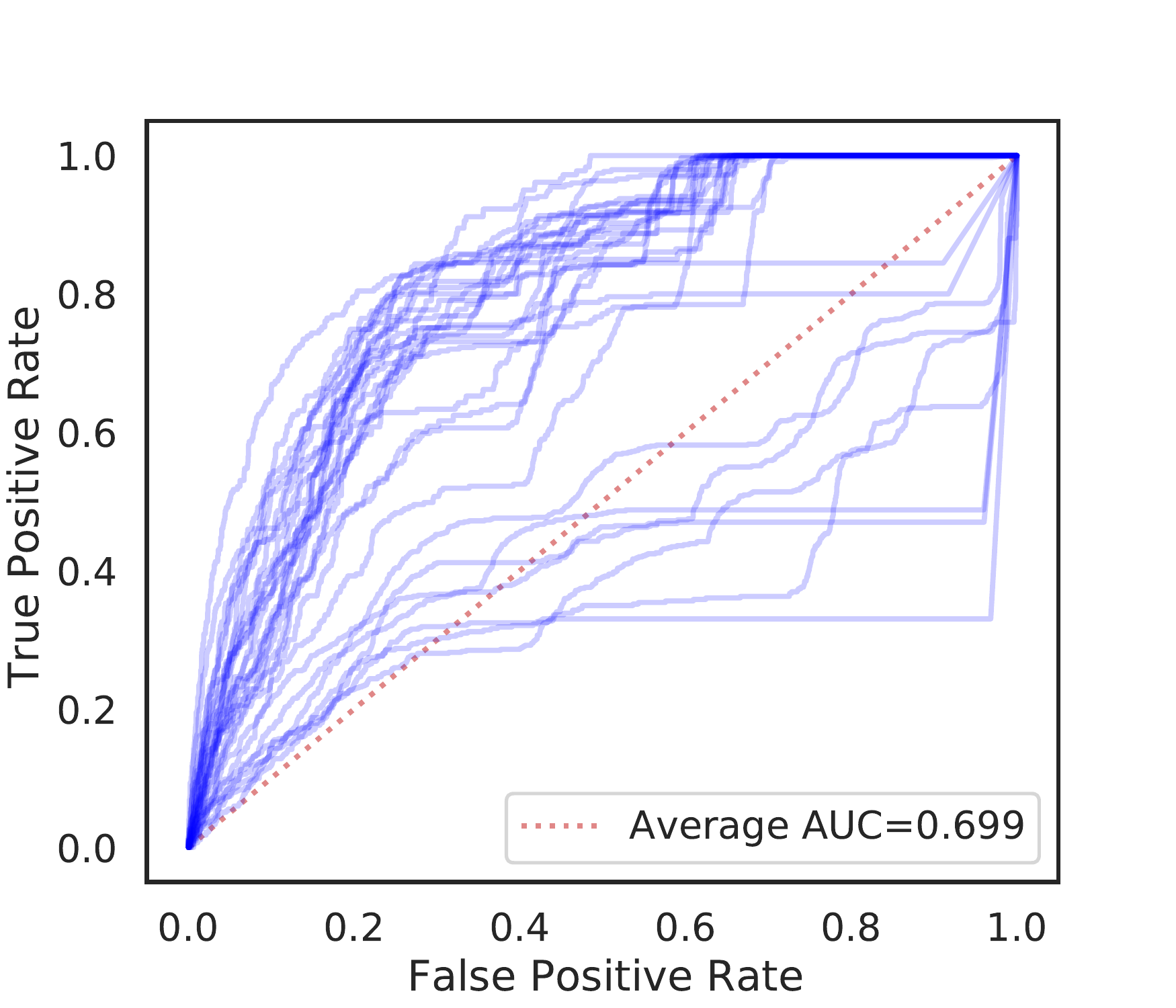}
    & \includegraphics[width=.5\textwidth]{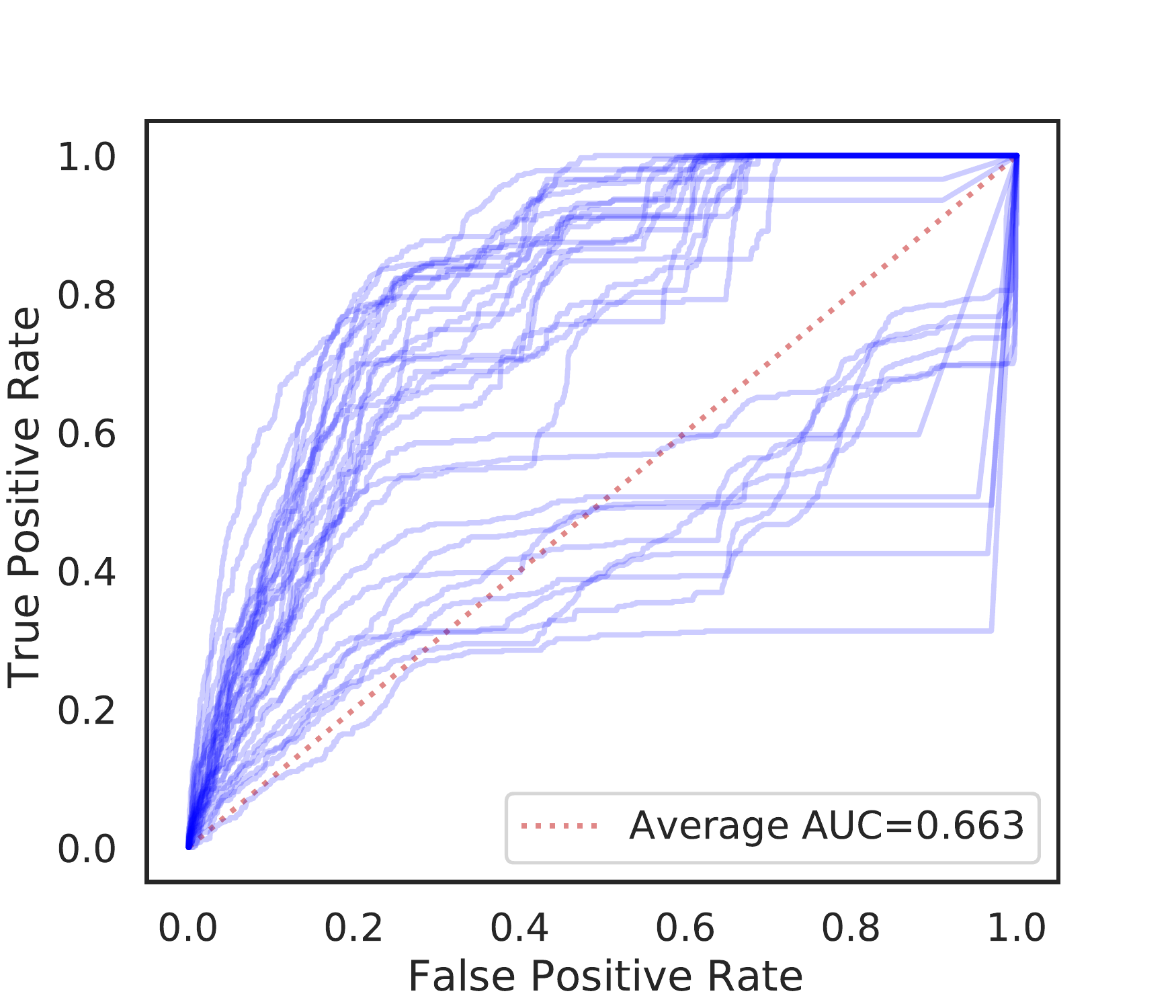} \\
    \textbf{\Large Adult (sex)}
    & \textbf{\Large Adult (race)} \\
    \includegraphics[width=.5\textwidth]{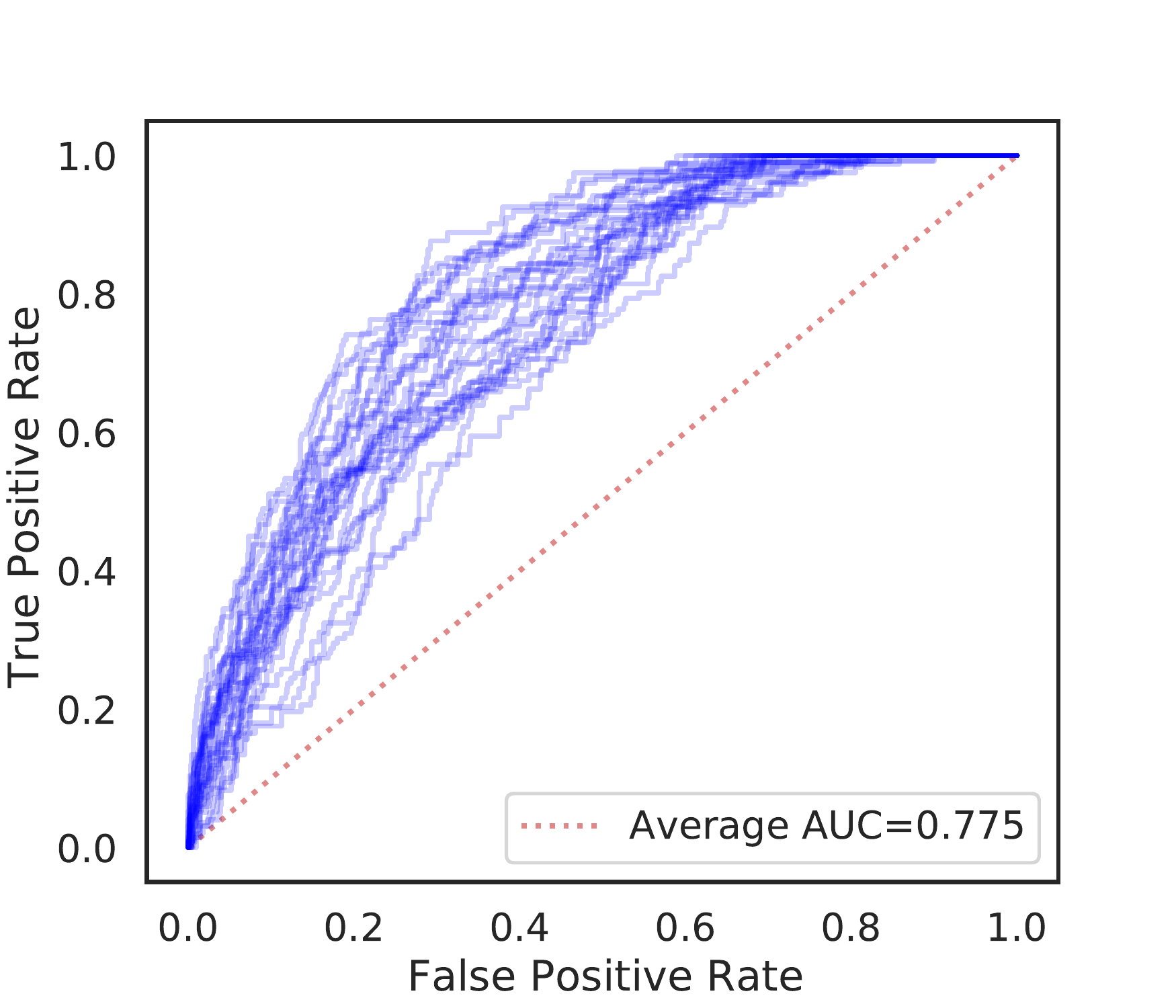}
    & \includegraphics[width=.5\textwidth]{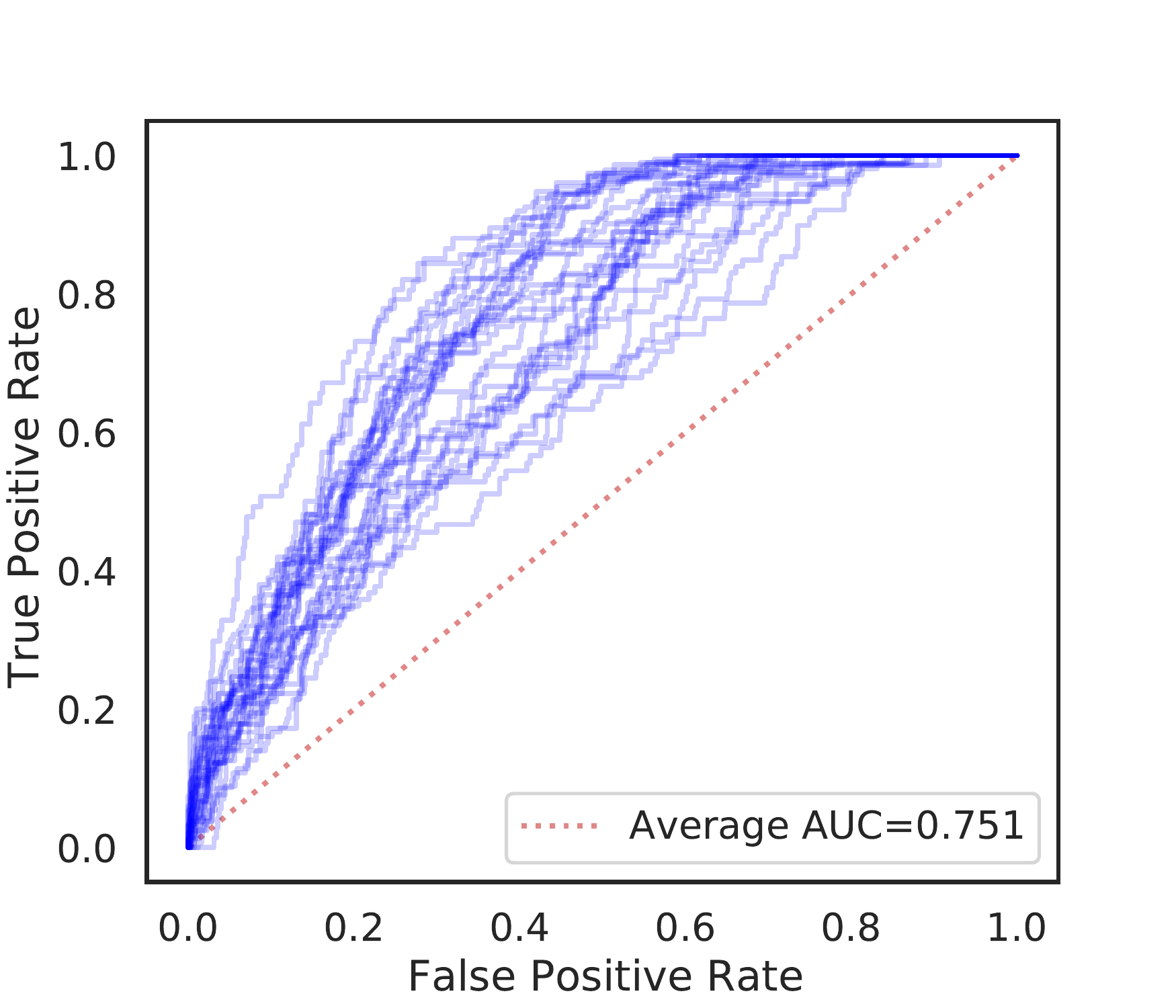} \\
    \textbf{\Large COMPAS (sex)}
    & \textbf{\Large COMPAS (race)} \\
  \end{tabular}
  \caption{Distributions of ROC curves for using prediction
    sensitivity to detect unfair predictions on real datasets, based
    on 30 trials for each configuration. In all cases, prediction
    sensitivity is highly predictive of non-membership in the match
    set, but its effectiveness varies with the data.}
  \label{fig:roc_curves}
\end{figure}

The results of our experiments on real data appear in
Figures~\ref{fig:group_fairness}, \ref{fig:mean_sensitivity},
and~\ref{fig:roc_curves}, and Table~\ref{tbl:full_numbers}; we
summarize them below.

\paragraph{Counterfactual augmentation improves group fairness metrics
  (Figure~\ref{fig:group_fairness}).}
Figure~\ref{fig:group_fairness} shows that as expected, classifiers
trained on counterfactually-augmented data score much better according
to common group fairness metrics than classifiers trained on the
original data. This result is expected, and consistent with previous
work~\cite{kaushik2019learning, kaushik2020explaining,
  black2020fliptest}.

\paragraph{Prediction sensitivity detects violations of counterfactual
  fairness (Figures~\ref{fig:mean_sensitivity}
  and~\ref{fig:roc_curves}).}
Figure~\ref{fig:mean_sensitivity} plots the distribution of prediction
sensitivities for members and non-members of the match sets
(``matching'' and ``non-matching,'' respectively). In both datasets,
and for both protected attributes, members of the match set have
consistently smaller prediction sensitivity; on average, non-members
have significantly higher values for prediction sensitivity.

Figure~\ref{fig:roc_curves} plots receiver operating characteristic
(ROC) curves and corresponding average area under the curve (AUC)
values for the distinguisher $\mathcal{D}$ built using prediction
sensitivity for each dataset. Each plot contains 30 ROC curves, one
per trial of the experiment. The results show that prediction
sensitivity is an effective approach for distinguishing between
members and non-members of the match set, and thus it is likely
effective at detecting failures of counterfactual fairness. The
results also demonstrate considerable variability, especially for the
Adult dataset. Both figures contain results based on a
counterfactually-augmented test set. Further discussion of these
results appears in Section~\ref{sec:discussion}.

\paragraph{Results are consistent for original and augmented test sets
  (Table~\ref{tbl:full_numbers}).}
As described in Section~\ref{sec:experiment-setup}, we performed each
experiment twice---once with the original test set, and once with a
counterfactually-augmented test set. Table~\ref{tbl:full_numbers}
compares the results of these two experiments. The results are nearly
identical, suggesting that prediction sensitivity is effective in both
settings. Table~\ref{tbl:full_numbers} also compares fully-trained
classifiers (40 epochs) against partially-trained classifiers (10
epochs). The impact of the fine-tuning that occurs in later stages of
training depends on the dataset; prediction sensitivity is more
effective for detecting violations in the fully-trained classifier for
the Adult dataset, but it is \emph{less} effective in this setting for
the COMPAS dataset.

\subsection{Discussion}
\label{sec:discussion}

The results described in Section~\ref{sec:exp2} suggest that
prediction sensitivity is an effective measure of counterfactual
fairness. However, the average area under the curve (AUC) for our
experiments on real datasets is significantly lower than the AUC for
our synthetic data experiments. We suspect two possible contributing
factors. First, it could be that bias is less severe in the real
datasets, so failures of counterfactual fairness are harder to detect.
This would also explain the difference in AUC between the two real
datasets. Second, our approach to counterfactual augmentation could be
missing significant correlations between other features and protected
status. This could cause low AUC by invalidating the match set
(prediction sensitivity could be \emph{correct}, while the match set
is \emph{wrong}). In particular, this factor may be responsible for
the plateaus in ROC curves for some trials on the Adult dataset
(Figure~\ref{fig:roc_curves}).

Our results suggest that prediction sensitivity is effective at
detecting unfair predictions, but they also reflect the inherent
challenge of this task. Individual predictions with extremely high
prediction sensitivity are likely to be blatantly unfair, and should
be easily detected using prediction sensitivity; however, borderline
cases may be more difficult to detect, as demonstrated by the
variability in the ROC curves in Figure~\ref{fig:roc_curves}.

\section{Conclusion}

We have presented a new, end-to-end approach for continual auditing of
counterfactual fairness in deployed deep learning systems.
We propose prediction sensitivity, an efficiently computed metric for
counterfactual fairness. Prediction sensitivity can be used at
prediction time, on deployed classifiers, to raise an alarm when the
classifier makes an unfair prediction. Prediction sensitivity handles
correlations between features and protected status, and does not
require access to individuals' protected status at prediction time.
Our empirical evaluation on synthetic and real datasets suggests that
prediction sensitivity is effective for detecting failures of
counterfactual fairness.

\iffinal
\begin{acks}
  We thank David Darais and Kristin Mills for their contributions to
  the development of this work. This research was supported in part by
  an Amazon Research Award.
\end{acks}
\fi

\bibliographystyle{plain}
\bibliography{refs,fairness}

\end{document}